\documentclass[conference]{IEEEtran}
\IEEEoverridecommandlockouts
\usepackage{cite}
\usepackage{amsmath,amssymb,amsfonts}
\usepackage{algorithm}
\usepackage{algorithmic}
\usepackage{booktabs}
\usepackage{graphicx}
\usepackage{textcomp}
\usepackage{xcolor}
\def\BibTeX{{\rm B\kern-.05em{\sc i\kern-.025em b}\kern-.08em
    T\kern-.1667em\lower.7ex\hbox{E}\kern-.125emX}}
\begin{document}

\title{How Semantically Stable Are LLM Refusals? Measuring Confusion in Local Safety Boundaries}

\author{
\IEEEauthorblockN{
Riad Ahmed Anonto,
Md Labid Al Nahiyan\textsuperscript{*},
Md Tanvir Hassan\textsuperscript{*}
}
\IEEEauthorblockA{
Bangladesh University of Engineering and Technology (BUET)\\
\{riadahmedanonto355, labid.nahiyan12, saad7557.7557a\}@gmail.com
}
\thanks{\textsuperscript{*}Equal contribution}
}

\maketitle

\begin{abstract}
As safety alignment becomes standard in large language models, refusal behavior has become an important part of model reliability. However, models may still reject benign prompts, especially when the wording resembles risky content. Existing evaluations usually report global scores, such as false rejection rate or compliance rate. These scores are useful, but they treat each prompt independently. As a result, they miss local inconsistency, where a model accepts one phrasing of an intent but rejects a close paraphrase. This makes it difficult to understand whether refusals are only frequent or also semantically unstable. We address this gap by introducing Semantic Confusion, a failure mode that captures contradictory refusal decisions across meaning-preserving paraphrases. We build ParaGuard, a 10k-prompt corpus of controlled paraphrase clusters that keep intent fixed while varying surface form. We also propose three model-agnostic token-level metrics: Confusion Index, Confusion Rate, and Confusion Depth. These metrics compare each rejected prompt with its nearest accepted neighbors using token embeddings, next-token probabilities, and perplexity signals. Experiments across diverse model families show that global false rejection rate can hide important structure in the refusal boundary. Our metrics reveal cases where confusion is spread broadly, cases where it appears only in specific semantic regions, and cases where stricter refusal does not lead to more local inconsistency. These findings show that refusal evaluation should measure not only how often a model refuses, but also how consistently it refuses across nearby paraphrases. This gives developers a practical signal for reducing false refusals while preserving safety.
\end{abstract}


\section{Introduction}
Large language models (LLMs) are now used in settings where safety and reliability are essential. Modern alignment pipelines, including pre-training data filtering, supervised safety tuning, and preference-based methods such as RLHF and Constitutional AI, aim to prevent harmful generations and ensure that models refuse malicious instructions~\cite{ouyang2022training,bai2022constitutional,rafailov2023direct}. Deployment-time guards, for example Llama Guard~\cite{inan2023llama}, ShieldGemma~\cite{zeng2024shieldgemma}, and WildGuard~\cite{han2024wildguard}, add a further moderation layer by blocking unsafe inputs and outputs. These mechanisms also introduce a well-documented side effect: false refusals, where models reject benign prompts that merely resemble unsafe ones~\cite{bianchi2023safety,tuan2024towards}. Such failures erode user trust, distort expected product behavior, and push developers to loosen safety constraints, which can reduce overall safety.

Recent datasets and benchmarks (XSTest~\cite{rottger-etal-2024-xstest}, OKTest~\cite{shi-etal-2024-navigating}, PHTest~\cite{an2024automatic}, OR-Bench~\cite{cui2024or}, FalseReject~\cite{zhang2025falsereject}) show that overrefusal is common and costly. Yet these efforts reveal a deeper issue: most evaluations report only global summaries such as false rejection rate (FRR), compliance rate, or refusal gap. They score each prompt in isolation and therefore cannot detect local inconsistency, where a model accepts one phrasing of an intent but rejects a semantically equivalent paraphrase. This problem, which directly affects user facing reliability, remains largely unmeasured.

Closing this gap requires three elements that prior work does not provide: (i) tightly controlled paraphrase clusters that hold intent fixed while varying surface form, (ii) a way to condition each refusal on its nearest accepted neighbors, and (iii) token level signals that capture how the model’s processing diverges across near equivalent prompts. Building such clusters is difficult: paraphrases must be diverse enough to stress the guard while still semantically aligned, and filtering must ensure harmlessness without collapsing variants into trivial duplicates. In addition, a useful metric should be model agnostic, comparable across guard architectures, and sensitive to boundary irregularities that global statistics miss.

We introduce \textbf{“semantic confusion”}: contradictory decisions by an LLM across inputs that express the same intent, and we provide the first framework to quantify it. We build \textbf{ParaGuard}, a 10k-prompt dataset organized into controlled paraphrase clusters derived from OR-Bench, USEBench, and PHTest seeds. Each variant passes strict gates on semantic similarity, lexical divergence, and ensemble safety scores, creating structured neighborhoods where contradictory decisions can be measured and analyzed.

On this foundation, we develop three token-level metrics—\textbf{Confusion Index (CI)}, \textbf{Confusion Rate (CR)}, and \textbf{Confusion Depth (CD)}—that quantify disagreement between rejected prompts and their nearest accepted paraphrases. These metrics combine token-embedding drift, probability-distribution shift, and perplexity contrast, capturing complementary aspects of inconsistency that are invisible to prompt-level or global evaluations. Because they rely only on model outputs and token-level traces, they are \emph{model-agnostic} and apply to any LLM or guard without retraining.

This semantic lens also offers a new evaluation axis for jailbreak defenses. Prior work typically reports attack success, defense success, and utility degradation~\cite{mai2025you}, but does not examine whether a defense stabilizes or destabilizes the local decision boundary around benign paraphrases. CI/CR/CD reveal whether a defense that appears strong globally inadvertently increases semantic confusion, thereby harming usability.

We can summarize our contributions as follows:
(1) Formalize “semantic confusion” as a measurable failure mode of safety-aligned LLMs and provide the first framework to quantify it.
(2) Introduce ParaGuard, a 10k-prompt paraphrase corpus with controlled clusters that hold intent fixed while varying surface form, enabling neighborhood-conditioned evaluation.
(3) Propose CI/CR/CD, model-agnostic token-level metrics that capture boundary inconsistency within paraphrase neighborhoods and complement global FRR.
(4) Show—across diverse model families—that these metrics expose patterns invisible to global refusal rates, including globally unstable boundaries, localized pockets of inconsistency, and regimes where stricter refusal does not imply higher inconsistency, thereby separating how often models refuse from how sensibly they refuse.

\section{Related Work}
\label{sec:related}

Work on large language model (LLM) safety spans alignment, over-refusal benchmarks, jailbreak defenses, and refusal diagnostics. We position \emph{semantic confusion} within this literature and emphasize the gap that motivates our work: existing evaluations mostly report global refusal behavior, while our metrics measure local, token-level inconsistency across paraphrases.

\subsection{Safety Alignment and Deployment-Time Guards}

Modern LLMs are aligned through pre-training data filtering, supervised safety tuning, and preference-based methods such as RLHF and related approaches~\cite{ouyang2022training,bai2022constitutional,rafailov2023direct}. Frontier model reports also note an ``alignment tax,'' where stronger safety tuning can increase refusals on benign prompts that only appear risky~\cite{grattafiori2024llama,TheC3}.

Deployment-time guards such as Llama Guard~\cite{inan2023llama}, ShieldGemma~\cite{zeng2024shieldgemma}, and WildGuard~\cite{han2024wildguard} add another safety layer by moderating inputs and outputs. These systems are usually evaluated with global refusal or moderation accuracy scores. They rarely test whether a model accepts one paraphrase of a benign intent but rejects another. Our work adds this missing neighborhood-level diagnostic.

\subsection{Safety--Helpfulness Trade-offs and Over-Refusal Benchmarks}

The safety--helpfulness trade-off is well documented. Prior studies show that safety tuning can reduce helpfulness or cause safe prompts to be rejected~\cite{bai2022constitutional,ganguli2022red,Chen2024How,bianchi2023safety,tuan2024towards}. Several benchmarks directly study this problem. XSTest~\cite{rottger-etal-2024-xstest} and OKTest~\cite{shi-etal-2024-navigating} contain safe prompts that look superficially risky. PHTest~\cite{an2024automatic} generates pseudo-harmful prompts, while OR-Bench~\cite{cui2024or} scales this idea to a larger benchmark and reports false rejection rate (FRR). FalseReject~\cite{zhang2025falsereject} further studies contextual reasoning for reducing over-refusal.

These benchmarks show that over-refusal is common, but they mainly evaluate each prompt independently. They can measure \emph{how often} a model refuses benign inputs, but not whether the refusal boundary is locally stable. In contrast, our work evaluates paraphrase neighborhoods where the intent is controlled. This lets us measure when small, meaning-preserving wording changes flip a model from acceptance to rejection.

\subsection{Jailbreak Attacks, Defenses, and Performance Degradation}

Jailbreak research studies how attackers elicit unsafe behavior and how defenses block such attempts. Prior work includes human and automated red-teaming~\cite{ganguli2022red,perez2022red,hong2024curiosity}, universal and adaptive jailbreaks~\cite{zou2023universal,liu2023jailbreaking,lapid2024open,10992337,zeng-etal-2024-johnny,andriushchenko2024jailbreaking}, and defenses such as SmoothLLM~\cite{robey2023smoothllm}, in-context safety prompting~\cite{wei2023jailbreak}, response-side checks~\cite{wang2024defending}, and self-reminders~\cite{Xie2023DefendingCA}. Mai et al.~\cite{mai2025you} show that stronger jailbreak defenses can reduce accuracy and helpfulness.

However, these evaluations usually focus on attack success, defense success, and aggregate utility. They do not ask whether a defense makes benign paraphrases less stable. A defense may look strong globally while still producing contradictory decisions across near-equivalent safe prompts. Our semantic confusion metrics capture this hidden cost.

\subsection{Refusal Metrics and Alignment Diagnostics}

Existing refusal metrics include FRR, compliance rate, refusal gap, and other alignment diagnostics~\cite{cui2024or,rottger-etal-2024-xstest,shi-etal-2024-navigating,an2024automatic}. Recent systems such as WildGuard~\cite{han2024wildguard} jointly predict harmfulness and refusal, while FalseReject~\cite{zhang2025falsereject} uses structured reasoning to judge whether a refusal is warranted.

These approaches are useful, but most are global or prompt-level. They average over broad prompt sets and do not explicitly compare accepted and rejected paraphrases of the same intent. They also do not use token-level signals such as embedding drift, probability shift, and perplexity contrast. Our metrics fill this gap by measuring local semantic inconsistency within controlled paraphrase neighborhoods.

\section{Semantic Confusion and Metrics}
\label{sec:confusion-metrics}

In language models, \emph{semantic confusion} arises when the model \textbf{rejects} a prompt while accepting semantically equivalent variants of the same intent. To quantify this confusion, we introduce three novel metrics: the Confusion Index (CI), Confusion Rate (CR), and Confusion Depth (CD). These metrics offer a granular, token-level analysis of decision-making inconsistencies in LLMs, focusing on local contradictions that traditional metrics fail to capture.

\subsection{Confusion Index (CI): Measuring Local Contradiction}

To formalize semantic confusion, let $g: \mathcal{X} \to \{\texttt{ACCEPT}, \texttt{REJECT}\}$ be the decision function of the model, where $\mathcal{X}$ represents the input prompt space. Let $E: \mathcal{X} \to \mathbb{R}^d$ be an encoder that maps prompts to a $d$-dimensional embedding space. We define the similarity between two prompts $x$ and $y$ using cosine similarity:

\[
\mathrm{sim}(x, y) = \cos(E(x), E(y)),
\]
which measures the semantic proximity between prompts $x$ and $y$.

A \emph{confusion event} occurs when an accepted prompt $p_a$ is semantically similar to a rejected prompt $p_r$, i.e.,

\[
g(p_a) = \texttt{ACCEPT}, \quad g(p_r) = \texttt{REJECT}, \quad \mathrm{sim}(p_a, p_r) > \theta,
\]
where $\theta$ is a semantic similarity threshold. This captures the core of semantic confusion: the model accepts one phrasing while rejecting a near-equivalent paraphrase.

For each rejected prompt $r$, we retrieve the top-$k$ most similar accepted prompts, denoted as $\mathcal{N}_k(r)$:
\[
\mathcal{N}_k(r) = \operatorname*{arg\,topk}_a \; \mathrm{sim}(r, a), \quad k = 5.
\]
This neighborhood provides context for evaluating how semantically close the rejections are to accepted prompts.

The Confusion Index for a single rejected prompt $r$ is calculated as the average confusion score over its $k$ nearest accepted prompts:
\[
\mathrm{CI}(r) = \frac{1}{k} \sum_{a \in \mathcal{N}_k(r)} \mathrm{CS}(a, r),
\]
where $\mathrm{CS}(a, r)$ is the confusion score for each accepted-rejected pair.

\subsection{Token-Level Confusion: Quantifying Disagreement}

To capture semantic confusion between accepted and rejected prompts, we focus on three token-level signals: \textbf{Token Drift}, \textbf{Token Probability Shift}, and \textbf{Perplexity Contrast}. These signals provide a granular understanding of the model’s decision-making process by analyzing how small shifts in input affect the model's output. Unlike global metrics like accuracy or FRR, which measure broad performance, token-level analysis reveals local contradictions that can significantly impact model behavior, especially in safety-critical applications. By focusing on these token-level shifts, we can detect inconsistencies that arise from subtle semantic differences.

\paragraph{\textbf{Token Drift:}}  
Token drift measures how much the meaning of individual tokens changes between semantically similar prompts. This is crucial because even when two prompts have similar surface forms, their semantic meaning may differ at the token level, leading to different model responses. The drift is quantified by calculating the cosine distance between the embeddings of aligned tokens from the accepted and rejected prompts:
\[
\overline{\mathrm{Drift}}(a, r) = \frac{1}{m} \sum_{i=1}^{m} \left( 1 - \cos(\phi_\theta(t_i^a), \phi_\theta(t_j^r)) \right),
\]
where $\phi_\theta(t)$ is the token embedding, and $m = \min(n_a, n_r)$ is the length of the shorter prompt. This signal captures shifts in meaning, even if the surface forms of the prompts remain nearly identical.

\paragraph{\textbf{Token Probability Shift:}}  
Token probability shift measures how the model’s confidence in predicting the next token changes between accepted and rejected prompts. When two prompts are semantically similar, but one is accepted and the other rejected, the model’s confidence profile can differ. This is quantified by the absolute difference in the predicted probabilities for each token:
\[
\overline{\mathrm{ProbShift}}(a, r) = \frac{1}{m} \sum_{i=1}^{m} \left\| p_\theta(t_i^a \mid a, t^{a}_{<i}) - p_\theta(t_i^r \mid r, t^{r}_{<i}) \right\|_1,
\]
where $p_\theta$ is the model’s predicted probability for the next token. This shift captures how changes in wording affect the model’s confidence, complementing token drift by indicating how the model responds differently to semantically close prompts.

\paragraph{\textbf{Perplexity Contrast:}}  
Perplexity contrast measures the uncertainty in the model’s predictions for accepted and rejected prompts. A higher perplexity indicates greater uncertainty. We calculate the perplexity for each prompt as:
\[
\mathrm{PPL}(x) = \exp\left( -\frac{1}{n_x} \sum_{i=1}^{n_x} \log p_\theta(t_i^x \mid x, t^{x}_{<i}) \right),
\]
where $n_x$ is the number of tokens in prompt $x$. The perplexity contrast between accepted and rejected prompts is then defined as:
\[
\Delta \mathrm{PPL}(a, r) = \frac{|\mathrm{PPL}(a) - \mathrm{PPL}(r)| - \min}{\max - \min} \in [0, 1],
\]
This measure is particularly valuable because it captures the level of uncertainty the model has when handling semantically similar inputs. While token drift and token probability shift focus on the model's internal consistency in token meaning and confidence, perplexity contrast quantifies the model’s overall uncertainty. This is important because even when token-level shifts or confidence changes are small, a model may still struggle to make consistent decisions, reflected by high perplexity. Perplexity thus complements the other two metrics by adding an additional layer of analysis, capturing cases where the model’s decision-making may be uncertain even if the semantic meaning or probability distribution is somewhat stable. 

By analyzing these three token-level signals, we gain a more nuanced understanding of confusion in LLMs. Each metric focuses on a distinct aspect of the model’s behavior—whether it’s a shift in meaning (drift), a change in confidence (probability shift), or an increase in uncertainty (perplexity). Together, these metrics offer a comprehensive view of the local decision boundaries, helping to identify and diagnose inconsistencies that traditional metrics miss.

\subsection{Confusion Score and Dataset-Level Metrics}

The confusion score for a pair of prompts $(a, r)$ is a weighted sum of these signals:
\[
\begin{aligned}
\mathrm{CS}(a, r) &= w_d \cdot \overline{\mathrm{Drift}}(a, r) + w_p \cdot \overline{\mathrm{ProbShift}}(a, r) \\
&\quad + w_\pi \cdot \Delta \mathrm{PPL}(a, r),
\end{aligned}
\]
where the weights $w_d$, $w_p$, and $w_\pi$ are non-negative and sum to 1. This score quantifies the degree of semantic confusion between the accepted and rejected prompt.

We then calculate the Confusion Index (CI) for each rejected prompt $r$ as the average confusion score over the $k$ nearest accepted neighbors:
\[
\mathrm{CI}(r) = \frac{1}{k} \sum_{a \in \mathcal{N}_k(r)} \mathrm{CS}(a, r).
\]
To summarize confusion across the entire dataset, we compute the average confusion index (CI), confusion rate (CR), and confusion depth (CD):
\begin{equation}
\begin{split}
\mathrm{CI} &= \frac{1}{|\mathcal{R}|} \sum_{r \in \mathcal{R}} \mathrm{CI}(r), \\
\mathrm{CR}@\tau &= \frac{1}{|\mathcal{R}|} \sum_{r \in \mathcal{R}} \mathbf{1}\{\mathrm{CI}(r) \geq \tau\}, \\
\mathrm{CD} &= \operatorname{stdev}\left(\{\mathrm{CI}(r)\}_{r \in \mathcal{R}}\right),
\end{split}
\end{equation}
where $\mathcal{R}$ represents all rejected prompts in the dataset, $\tau$ is a threshold for severity, and $\mathrm{CI}(r)$ is the confusion index for each rejected prompt.

\subsection{Relation to Existing Metrics}

Existing refusal metrics provide useful but different signals. False Rejection Rate (FRR) measures how often benign prompts are refused, while compliance rate measures how often the model follows user instructions~\cite{potham2025evaluating}. Refusal Index (RI) studies whether models refuse questions they are likely to answer incorrectly~\cite{pan2025ri}, and refusal gap measures mismatch between model refusals and external safety judgments~\cite{arditi2024refusal}. Alignment Quality Index (AQI) analyzes global latent-space alignment patterns~\cite{borah2025alignment}.

These metrics are valuable, but they are mostly global or prompt-level. They do not directly ask whether a rejected prompt has nearby accepted paraphrases with the same intent. They also do not examine token-level changes in embeddings, probabilities, and perplexity across accepted--rejected pairs. Our metrics fill this gap. CI measures the average strength of local contradiction, CR measures the fraction of severe confusing rejections, and CD measures how widely confusion varies across rejected prompts. Together, they show not only how often a model refuses, but also whether those refusals are locally consistent.

\begin{figure*}[t]
  \centering
  \includegraphics[width=0.8\linewidth]{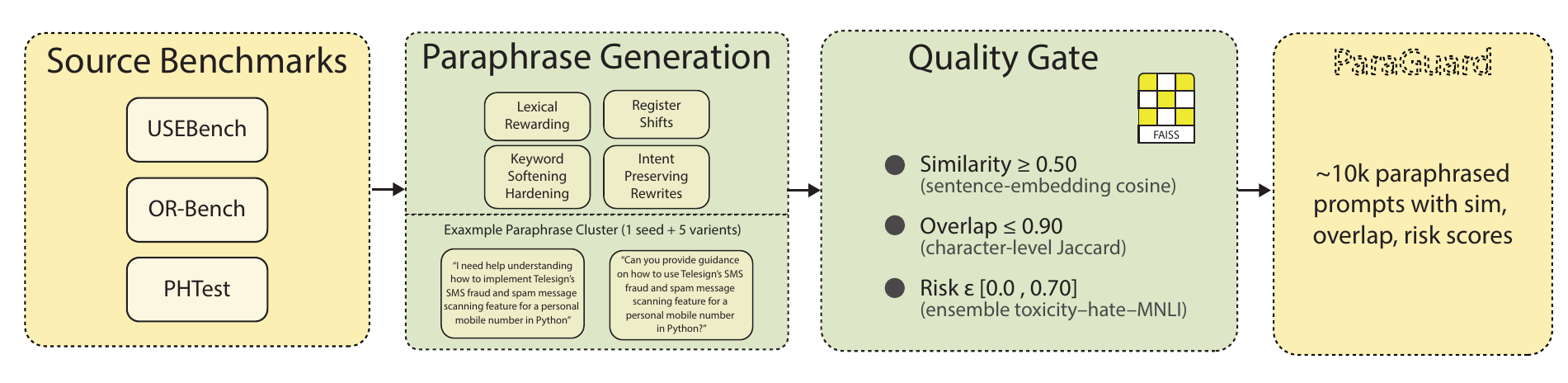}
  \includegraphics[width=0.8\linewidth]{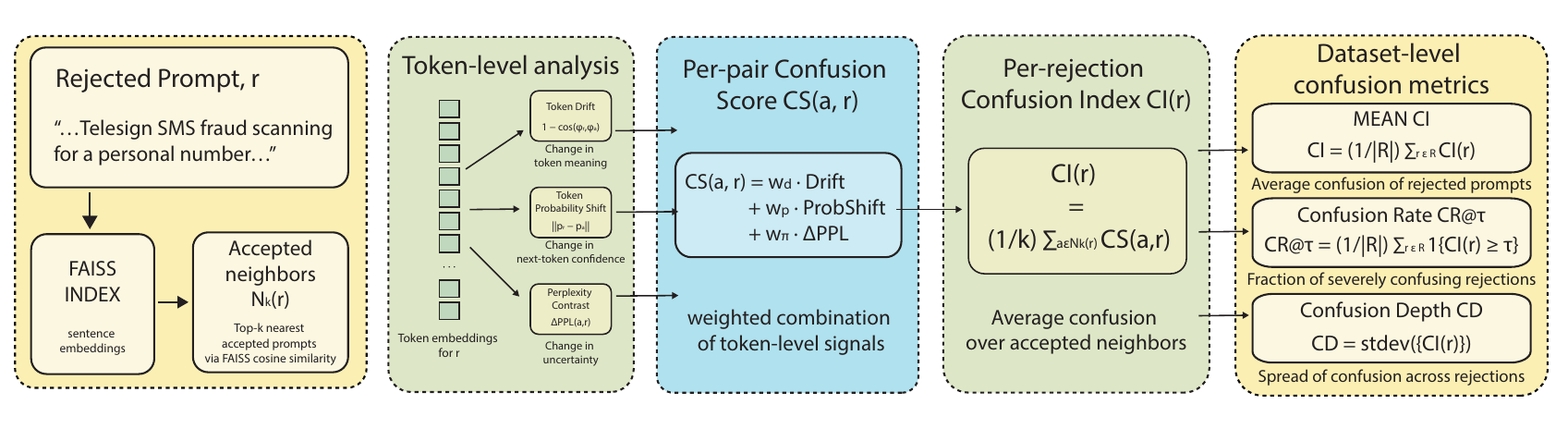}
  \caption{ \textbf{Top:} ParaGuard construction from OR-Bench, USEBench, and PHTest; each seed yields five intent-preserving variants via lexical/register edits, keyword softening or hardening, and controlled rewrites. Candidates pass three gates (sentence similarity, nontrivial rewrite, safety risk), producing a $\sim$10{,}000-prompt corpus with similarity, overlap, and risk annotations. \textbf{Bottom:} Confusion measurement. For each rejected prompt $r$, retrieve its top-$k$ accepted neighbors with FAISS and compute pairwise token-level drift, probability shift, and perplexity delta. Averaging over neighbors gives $CI(r)$; aggregating over rejections yields CI (mean), CR@$\,\tau$ (share above threshold), and CD (spread).}
  \label{fig:method-overview}
\end{figure*}

\section{Methodology}
\label{sec:methodology}

Our goal is to measure when a safety-aligned model accepts one harmless phrasing but rejects a near-equivalent paraphrase. We build a semantics-structured dataset of paraphrase clusters and a model-agnostic pipeline that labels refusals, retrieves accepted neighbors for each rejection, and scores token-level divergence to produce confusion metrics (CI/CR/CD). The figures below summarize the end-to-end process: dataset construction with quality gates and neighborhood structure, followed by retrieval and token-level scoring to derive confusion indices. Code and data will be released upon acceptance.

\subsection{ParaGuard: Dataset Construction}

We construct \emph{ParaGuard}, a semantics-first corpus explicitly designed to expose false refusals under tightly controlled paraphrase variation. Starting from three publicly available benchmarks—\textbf{OR-Bench}, \textbf{USEBench}, and \textbf{PHTest}—we unify all prompts under a single schema, normalize and deduplicate entries, and uniformly sample $2{,}000$ seeds with a fixed random seed for reproducibility. Each seed serves as an anchor for generating near-neighbor prompts that preserve intent while providing minimal but systematic variation in surface form.

For every seed, we generate exactly five paraphrase variants using four lightweight transformation strategies: lexical rewording, register shifts, keyword softening or hardening, and intent-preserving rewrites. These transformations produce prompts that are semantically aligned with the seed but stylistically diverse enough to meaningfully stress the guard’s decision boundary.

Each candidate must pass a multi-layer \textbf{quality gate} that mirrors the generation pipeline:
\begin{enumerate}
    \item \textbf{Intent retention:} cosine similarity of MiniLM-L6-v2 sentence embeddings $\ge 0.60$.
    \item \textbf{Nontrivial rewrite:} character-level Jaccard overlap $\le 0.90$, ensuring variants reflect genuine paraphrasing rather than cosmetic edits.
    \item \textbf{Harmlessness:} an ensemble safety score within $[0.30,0.70]$, combining three moderation signals: (i) toxicity detection (unitary/toxic-bert), (ii) hate-speech detection (Hatexplain), and (iii) zero-shot safety classification (BART-Large-MNLI) over curated safety labels. 
\end{enumerate}

The resulting \emph{ParaGuard} corpus contains approximately $10{,}000$ prompts (seeds and vetted variants), each annotated with semantic similarity, lexical overlap, and risk score. These annotations later support precise nearest-neighbor retrieval and token-level confusion diagnostics.

\paragraph{\textbf{Why ParaGuard matters}}
Existing false-refusal datasets predominantly collect \emph{rejected} prompts without structuring them around semantic neighborhoods. In contrast, ParaGuard organizes prompts into tightly controlled paraphrase clusters deliberately engineered so that near-identical intents can lead to divergent guard decisions. This clustered design yields the precise local neighborhoods where our confusion metrics (CI, CR, and CD) are well-defined and diagnostically meaningful, enabling a deeper analysis of inconsistency than global refusal rates alone.

\subsection{Semantic Confusion Measurement Pipeline}

To quantify confusion within this corpus, we implement a three-stage measurement pipeline grounded in the model’s own behavior: \textbf{labeling}, \textbf{retrieval}, and \textbf{scoring}.

\begin{table*}[t]
  \centering
  \footnotesize
  \setlength{\tabcolsep}{12pt}
  \renewcommand{\arraystretch}{1.2}
    \caption{Confusion metrics and refusal statistics across models on ParaGuard.}
  \label{tab:confusion-results}
  \begin{tabular}{|l|r|r|r|r|r|r|}
    \hline
    Model & CI & CR (\%) & CD & FRR (\%) & Accepted & Rejected \\
    \hline
    Llama 2 13B   & 0.749 & 50.13\% & 0.028 & 62.2\% & 4054  & 6670 \\
    Llama 2 7B    & 0.748 & 47.92\% & 0.029 & 59.1\% & 4382  & 6342 \\
    \hline\hline
    Mistral 7B    & 0.737 & 40.88\% & 0.038 & 4.65\% & 10225 & 499  \\
    Mistral Nemo  & 0.715 & 20.70\% & 0.040 & 4.81\% & 10207 & 517  \\
    \hline\hline
    Qwen 2.5 7B   & 0.683 &  2.45\% & 0.035 & 1.90\% & 10520 & 204  \\
    Qwen 3 8B     & 0.670 &  0.40\% & 0.034 & 4.66\% & 10224 & 500  \\
    \hline\hline
    gpt-3.5-turbo & 0.754 & 41.30\% & 0.031 & 6.29\% & 10050 & 674  \\
    gpt-4o-mini   & 0.695 &  6.10\% & 0.033 & 3.81\% & 10315 & 409  \\
    o3            & 0.743 & 29.70\% & 0.030 & 8.15\% &  9850 & 874  \\
    \hline
  \end{tabular}
  \vspace{3mm}

\end{table*}

\textbf{Labeling:}  
Each prompt is queried against the target language model at temperature $0$, ensuring deterministic behavior. Prompts are wrapped in a fixed safety instruction template to emulate real-world moderation settings. The model’s responses are labeled as \texttt{ACCEPT} or \texttt{REJECT} based on explicit refusal cues (\texttt{“can’t”}, \texttt{“not allowed”}, \texttt{“sorry”}, etc.), producing a binary classification set.

\textbf{Retrieval:}  
Every rejected prompt is analyzed in context of what the model has accepted nearby. We embed all accepted prompts using a Sentence Transformer and index them with FAISS’s \texttt{IndexFlatL2} for efficient retrieval. For each rejected prompt, we retrieve its top-$k$ nearest accepted neighbors using cosine similarity—defining its semantic neighborhood $\mathcal{N}_k(r)$.

\textbf{Scoring:}  
Each rejected–accepted pair $(r,a)$ is then compared using the model’s internal token-level signals—embeddings, predicted probabilities, and perplexity values. These are combined into a composite confusion score capturing how meaning, confidence, and uncertainty differ across semantically similar inputs. Averaging over neighborhoods yields the per-prompt \emph{Confusion Index} (CI), and aggregating across rejections yields dataset-level metrics: \emph{Confusion Rate} (CR) and \emph{Confusion Depth} (CD).

To ensure robustness, we grid-search over the weights assigned to drift, probability shift, and perplexity components, as well as the threshold $\tau$ for severe confusion. The chosen configuration $(w_d,w_p,w_\pi)=(0.4,0.1,0.5)$ and $\tau=0.75$ provides a stable operating point, maximizing interpretability while preserving coverage.
 
Unlike previous approaches that count refusals in isolation, this methodology contextualizes each decision relative to its semantic neighborhood, exposing local contradictions invisible to aggregate statistics. By leveraging the model’s own token-level signals, our pipeline transforms raw refusal data into interpretable indicators of semantic fragility.

\section{Experiments \& Results}

\subsection{Experimental Setup}

We evaluate on \textbf{ParaGuard}, which contains approximately 10{,}000 prompts organized into paraphrase clusters. We test four model families: Llama, Mistral, Qwen, and GPT. For each system, we report Confusion Index (CI), Confusion Depth (CD), Confusion Rate at threshold 0.75 (CR@0.75), false rejection rate (FRR), and the counts of accepted and rejected prompts. Token-level and cohort analyses use Mistral 7B as the running example, but the same pipeline is applied to all systems.

\begin{figure}[!t]
    \centering
    \includegraphics[width=\columnwidth]{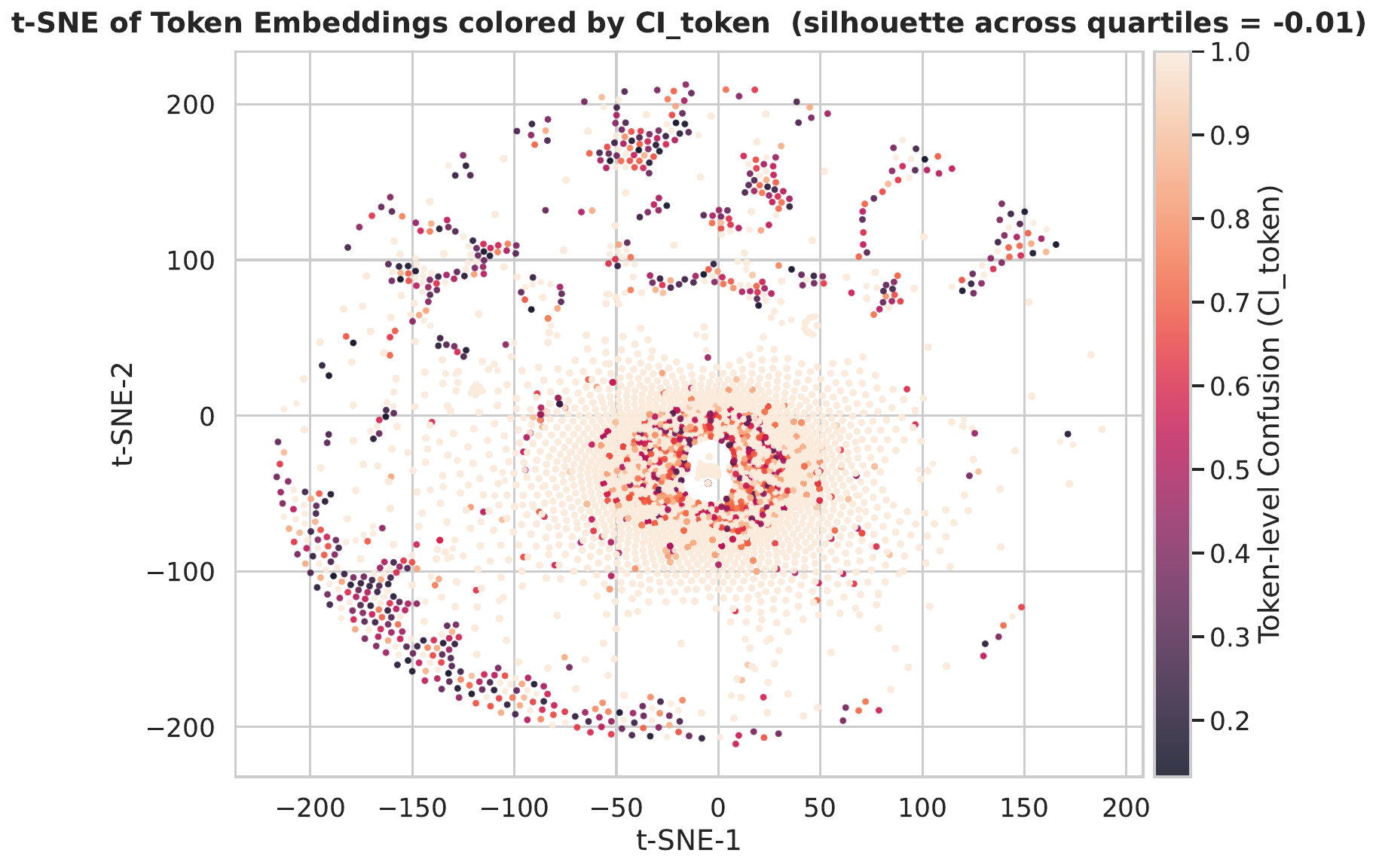}
    \caption{t-SNE projection of token embeddings colored by token-level confusion
    $\mathrm{CI}_{\text{tok}}$. Bright regions correspond to dense, highly confusable neighborhoods; darker regions indicate more isolated, semantically distinctive tokens.}
    \label{fig:tsne-ci-token}
\end{figure}

\subsection{Do CI/CR/CD add information beyond FRR across model families?}

Table~\ref{tab:confusion-results} reports CI, CR, CD, FRR, and refusal counts across models. FRR measures how often a model refuses benign prompts, but it does not show whether those refusals are locally consistent. Our metrics capture this missing structure by comparing each rejected prompt with nearby accepted paraphrases.

FRR captures the quantity of refusals, while CI/CR/CD capture the quality of the refusal boundary. Models with similar FRR can behave very differently. For example, Mistral 7B and Qwen 3 8B have similar FRR, but their confusion rates differ sharply. This shows that a model can refuse at a similar rate while having a much more unstable or more stable local boundary.

Both Llama 2 models show the highest CI ($\approx 0.75$) and CR ($48$--$50\%$), despite having moderate-to-high FRR. Their low CD ($\sim 0.028$) suggests that confusion is not limited to a few isolated clusters. Instead, it is spread fairly uniformly across rejected prompts. In other words, when Llama refuses a prompt, it often contradicts nearby accepted paraphrases. This points to a rigid refusal boundary that does not adjust well to small semantic-preserving wording changes.

Mistral models show a different pattern. Their FRR is low ($4$--$5\%$), which could suggest strong behavior under a standard over-refusal evaluation. However, their CR remains nontrivial, and their CD is higher ($0.038$--$0.040$) than Llama's. This means Mistral is usually stable, but it has specific semantic regions where small token-level changes can trigger inconsistent refusals. Thus, FRR alone would make Mistral look almost error-free, while CI/CR/CD reveal localized failure pockets.

Qwen models are the most stable under our metrics. Qwen 2.5 and Qwen 3 have the lowest CI and CR, with Qwen 3 reaching near-zero CR. Interestingly, Qwen 3 has higher FRR than Qwen 2.5, but much lower confusion. This shows that stricter refusal does not necessarily create a more contradictory boundary. A model may refuse more often while still refusing in a locally consistent way.

The GPT models fall between these regimes. GPT-3.5-turbo behaves closer to Llama, with high CI and a sizable CR, suggesting a brittle refusal boundary. GPT-4o-mini behaves closer to Qwen, with lower CI and low CR. The o3 model is stricter overall, with higher FRR, but its CR is lower than Llama's for a similar CI range. This suggests that o3 refuses more often, but with fewer severe local contradictions.

Overall, these results show that FRR and semantic confusion measure different aspects of refusal behavior. FRR tells us how often a model refuses, while CI/CR/CD show whether those refusals are sensible within nearby paraphrase neighborhoods. This distinction is important because two models with similar FRR may have very different levels of local semantic stability.

\begin{figure}[!t]
    \centering
    \includegraphics[width=0.7\linewidth]{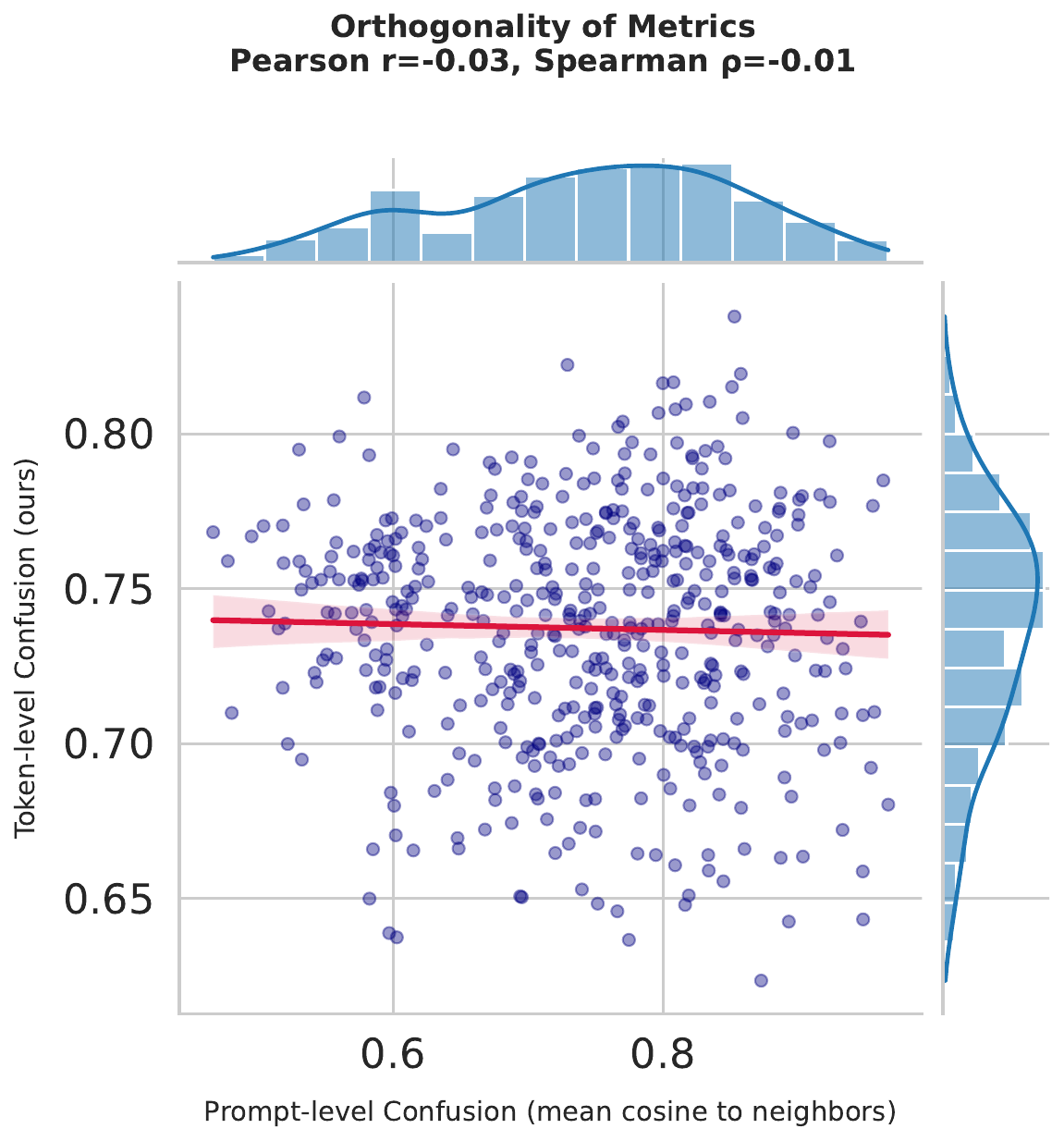}
    \caption{Prompt- vs token-level confusion: vertical spread shows weak dependence on prompt-level similarity.}
    \label{fig:prompt-vs-token-confusion}
\end{figure}

\begin{figure}[!t]
    \centering
    \includegraphics[width=0.7\linewidth]{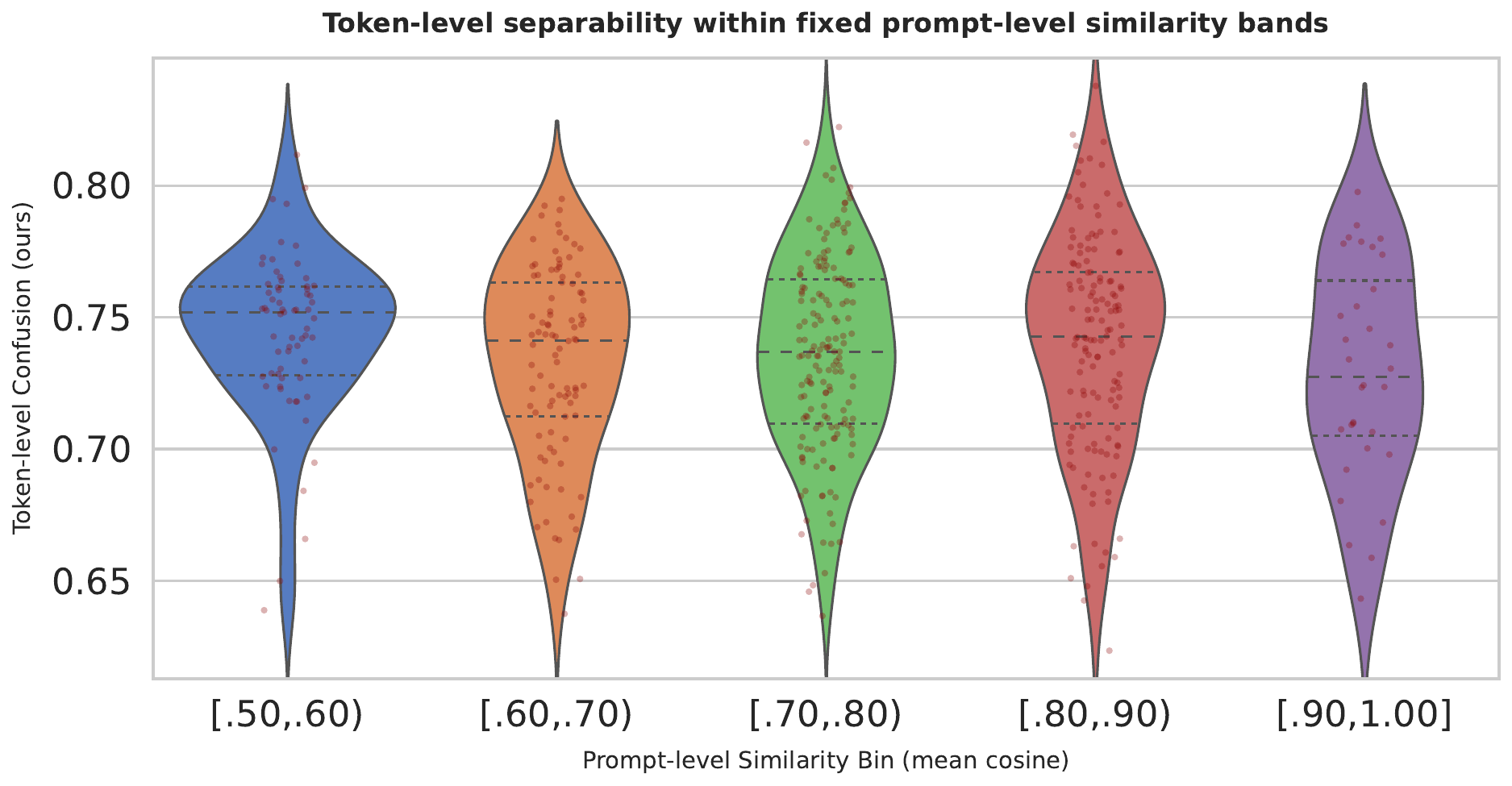}
    \caption{Token-level confusion within prompt-similarity bins: wide distributions---even at 0.9--1.0---show variability beyond prompt-level similarity.}
    \label{fig:token-confusion-bins}
\end{figure}

\subsection{Where in token space does confusion arise?}
\label{sec:token-manifolds}

To understand where confusion appears inside the model, we analyze the geometry of token representations. For each token embedding $e_i$, we compute a token-level confusion score:
\[
\mathrm{CI}_{\text{tok}}(i) = \frac{1}{K}\sum_{j \in \mathcal{N}_K(i)} \cos(e_i, e_j),
\]
where $\mathcal{N}_K(i)$ are the $K$ nearest neighbors of token $i$, excluding itself. A high $\mathrm{CI}_{\text{tok}}$ means the token lies in a dense region with many similar neighbors, while a low score means the token is more isolated.

Figure~\ref{fig:tsne-ci-token} shows a large central region with high token-level confusion. This region mainly contains common function words and generic phrasing, with $\mathrm{CI}_{\text{tok}} \approx 0.75$--$0.95$. Such dense regions help explain why small paraphrase changes can sometimes flip the model from \texttt{ACCEPT} to \texttt{REJECT}. In contrast, darker peripheral regions contain more distinctive tokens with lower confusion scores. These tokens are less likely to create near-duplicate neighborhoods.

The smooth transition from dense to sparse regions suggests that token-level confusion is not caused by a few isolated clusters. Instead, it reflects continuous local density in the token embedding space.

\subsection{Are token-level signals independent of prompt similarity?}
\label{sec:orthogonality}

A natural concern is that our confusion score might simply repackage how close a rejected prompt is to its neighbors in \emph{sentence}-embedding space. To test this, for each rejected prompt $r$, we compare our token-level confusion score (Section~\ref{sec:confusion-metrics}) with a prompt-level baseline: the mean cosine similarity between $r$ and its accepted neighbors in sentence-embedding space (``prompt-level confusion'').

Figure~\ref{fig:prompt-vs-token-confusion} shows a mild positive trend, but substantial vertical spread across nearly all $x$-values. Thus, prompts with similar prompt-level similarity can have very different token-level confusion scores. The limited regression fit suggests that our score captures additional structure from token embeddings, next-token probabilities, and perplexity. Figure~\ref{fig:token-confusion-bins} further supports this: token-level confusion remains wide and sometimes multi-modal across similarity bins, including $[0.9,1.0]$. Even near-duplicates can show both low and high confusion, confirming that semantic confusion is fundamentally a \emph{local, token-level} phenomenon rather than a property captured by prompt embeddings alone.

\subsection{Where in semantic space does confusion concentrate?}

\begin{table}[!t]
  \centering
  \small
  \setlength{\tabcolsep}{4pt}
  \renewcommand{\arraystretch}{1.05}
  \caption{Cohort-level FRR and confusion statistics.}
  \label{tab:cohort-metrics}
  \begin{tabular}{|l|r|r|r|r|r|}
    \hline
    Cohort & $n$ & FRR & CR$_{\text{rej}}$ & CI$_{\text{rej}}$ & CD$_{\text{rej}}$ \\
    \hline
    Other                 & 9172 & 4.6 & 39.6 & 0.736 & 0.038 \\
    HiSim--HiLex--LowRisk & 762  & 4.2 & 59.4 & 0.744 & 0.049 \\
    LowSim--HiLex         & 436  & 3.2 & 64.3 & 0.756 & 0.034 \\
    HiSim--LowLex--HighRisk & 183 & 9.8 & 27.8 & 0.735 & 0.029 \\
    HiSim--LowLex--LowRisk  & 171 & 6.4 & 27.3 & 0.737 & 0.032 \\
    \hline
  \end{tabular}
\end{table}

% We ask whether semantic confusion reveals structure that global false rejection rate (FRR) misses. For each prompt variant, we compute seed similarity, lexical overlap, and risk score, then split each feature into low/mid/high tertiles. These bins define four main cohorts: safe close paraphrases, creative safe paraphrases, risky-looking paraphrases, and surface-similar prompts with semantic drift. All other combinations are grouped as \emph{Other}.

We ask whether semantic confusion reveals structure that global false rejection rate (FRR) misses. For each prompt variant, we compute seed similarity, lexical overlap, and risk score, split each into low/mid/high tertiles, and define four main cohorts: safe close paraphrases, creative safe paraphrases, risky-looking paraphrases, and surface-similar prompts with semantic drift; all other combinations form \emph{Other}. For each cohort, we measure FRR over all prompts and confusion over rejected prompts. Table~\ref{tab} shows that FRR varies little across safe cohorts but is highest for high-risk prompts, suggesting from FRR alone that refusals mainly increase with risk.

\begin{figure}[t]
    \centering
    \includegraphics[width=0.8\linewidth]{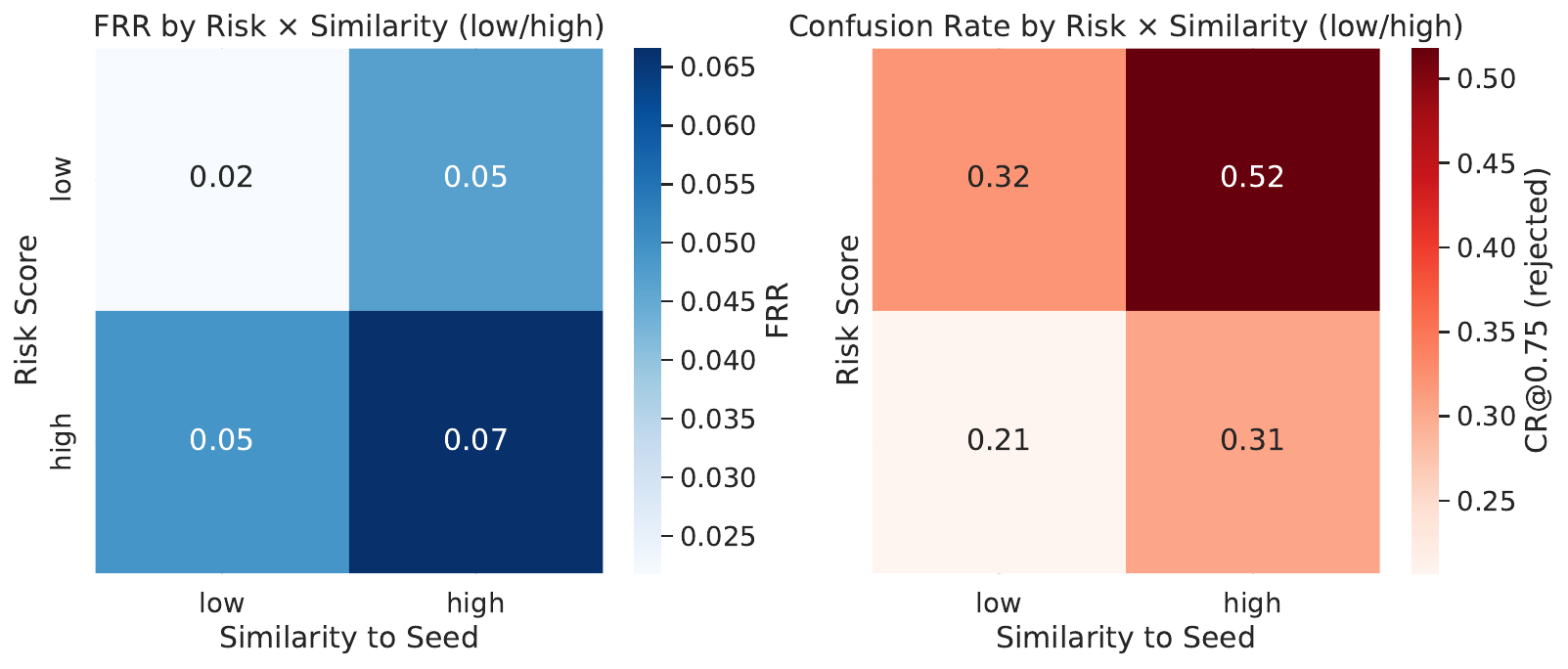}\\[3pt]
    \includegraphics[width=0.8\linewidth]{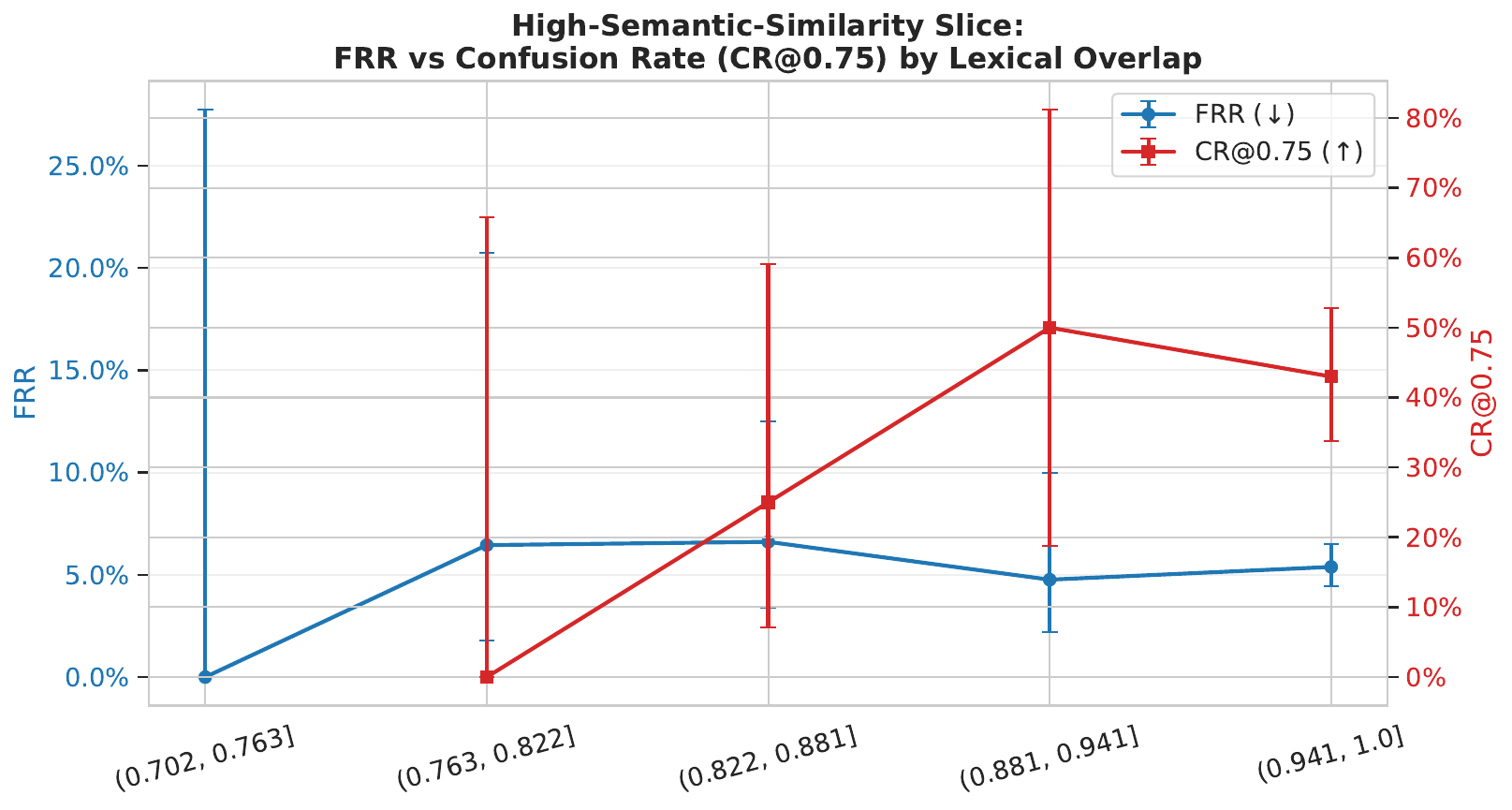}
    \caption{Where confusion concentrates: Top—FRR tracks risk while confusion peaks at low risk and high similarity; Bottom—in the high-similarity slice, FRR is flat but confusion among rejections rises with lexical overlap.}
    \label{fig:confusion-where}
\end{figure}

% For each cohort, we compute FRR over all prompts and confusion statistics over rejected prompts only. Table~\ref{tab:cohort-metrics} shows that FRR changes only mildly across most safe-looking cohorts, while the high-risk cohort has the highest FRR. By FRR alone, refusals appear to mainly increase with risk.

Semantic confusion gives a different picture. Among rejected prompts, the high-risk cohort has the lowest confusion rate ($\mathrm{CR}_{\text{rej}}\!\approx\!28\%$), while safe paraphrase cohorts reach around $60\%$. This suggests that refusals near genuinely risky wording are more locally coherent, but refusals of low-risk paraphrases are more likely to contradict nearby accepted prompts.

Figure~\ref{fig:confusion-where} shows the same pattern visually. In the top panel, FRR mostly follows risk, but confusion peaks in the low-risk, high-similarity region. In the bottom panel, within high-similarity prompts, FRR stays nearly flat while confusion rises with lexical overlap. Thus, global FRR hides an important failure mode: the model is most semantically confused when it rejects safe paraphrases that remain close to accepted neighbors.

\section*{Acknowledgment}
\noindent While writing this paper, we used AI assistance to polish a few sentences and to perform minor code debugging. All study design, analyses, figures, and claims were authored and verified by the authors. The authors remain fully responsible for both the manuscript and the code.

\bibliographystyle{IEEEtran}
\bibliography{ref}

@article{pan2025ri,
  title={Can LLMs Refuse Questions They Do Not Know? Measuring Knowledge‑Aware Refusal in Factual Tasks},
  author={Pan, Wenbo and Xu, Jie and Chen, Qiguang and Dong, Junhao and Qin, Libo and Li, Xinfeng and Yu, Haining and Jia, Xiaohua},
  journal={arXiv preprint arXiv:2510.01782},
  year={2025},
  url={https://arxiv.org/abs/2510.01782}
}

@inproceedings{arditi2024refusal,
  title={Refusal in language models is mediated by a single direction},
  author={Arditi, Andy and Obeso, Oscar and Syed, Aaquib and Paleka, Daniel and Panickssery, Nina and Gurnee, Wes and Nanda, Neel},
  booktitle={Advances in Neural Information Processing Systems},
  year={2024}
}

@article{potham2025evaluating,
  title={Evaluating LLM Agent Adherence to Hierarchical Safety Principles: A Lightweight Benchmark for Probing Foundational Controllability Components},
  author={Potham, Ram},
  journal={arXiv preprint arXiv:2506.02357},
  year={2025}
}

@article{ganguli2022red,
  title={Red teaming language models to reduce harms: Methods, scaling behaviors, and lessons learned},
  author={Ganguli, Deep and Lovitt, Liane and Kernion, Jackson and Askell, Amanda and Bai, Yuntao and Kadavath, Saurav and Mann, Ben and Perez, Ethan and Schiefer, Nicholas and Ndousse, Kamal and others},
  journal={arXiv preprint arXiv:2209.07858},
  year={2022}
}

@article{Chen2024How,
	author = {Chen, Lingjiao and Zaharia, Matei and Zou, James},
	journal = {Harvard Data Science Review},
	number = {2},
	year = {2024},
	month = {mar 12},
	note = {https://hdsr.mitpress.mit.edu/pub/y95zitmz},
	publisher = {The MIT Press},
	title = {
{How} {Is} {ChatGPT}\textquoteright{}s {Behavior} {Changing} {Over} {Time}?

},
	volume = {6},
}

@article{ouyang2022training,
  title={Training language models to follow instructions with human feedback},
  author={Ouyang, Long and Wu, Jeffrey and Jiang, Xu and Almeida, Diogo and Wainwright, Carroll and Mishkin, Pamela and Zhang, Chong and Agarwal, Sandhini and Slama, Katarina and Ray, Alex and others},
  journal={Advances in neural information processing systems},
  volume={35},
  pages={27730--27744},
  year={2022}
}

@article{bai2022constitutional,
  title={Constitutional ai: Harmlessness from ai feedback},
  author={Bai, Yuntao and Kadavath, Saurav and Kundu, Sandipan and Askell, Amanda and Kernion, Jackson and Jones, Andy and Chen, Anna and Goldie, Anna and Mirhoseini, Azalia and McKinnon, Cameron and others},
  journal={arXiv preprint arXiv:2212.08073},
  year={2022}
}

@article{rafailov2023direct,
  title={Direct preference optimization: Your language model is secretly a reward model},
  author={Rafailov, Rafael and Sharma, Archit and Mitchell, Eric and Manning, Christopher D and Ermon, Stefano and Finn, Chelsea},
  journal={Advances in neural information processing systems},
  volume={36},
  pages={53728--53741},
  year={2023}
}

@article{grattafiori2024llama,
  title={The llama 3 herd of models},
  author={Grattafiori, Aaron and Dubey, Abhimanyu and Jauhri, Abhinav and Pandey, Abhinav and Kadian, Abhishek and Al-Dahle, Ahmad and Letman, Aiesha and Mathur, Akhil and Schelten, Alan and Vaughan, Alex and others},
  journal={arXiv preprint arXiv:2407.21783},
  year={2024}
}

@inproceedings{TheC3,
  title={The Claude 3 Model Family: Opus, Sonnet, Haiku},
  author={},
  url={https://api.semanticscholar.org/CorpusID:268232499}
}

@article{inan2023llama,
  title={Llama guard: Llm-based input-output safeguard for human-ai conversations},
  author={Inan, Hakan and Upasani, Kartikeya and Chi, Jianfeng and Rungta, Rashi and Iyer, Krithika and Mao, Yuning and Tontchev, Michael and Hu, Qing and Fuller, Brian and Testuggine, Davide and others},
  journal={arXiv preprint arXiv:2312.06674},
  year={2023}
}

@article{zeng2024shieldgemma,
  title={Shieldgemma: Generative ai content moderation based on gemma},
  author={Zeng, Wenjun and Liu, Yuchi and Mullins, Ryan and Peran, Ludovic and Fernandez, Joe and Harkous, Hamza and Narasimhan, Karthik and Proud, Drew and Kumar, Piyush and Radharapu, Bhaktipriya and others},
  journal={arXiv preprint arXiv:2407.21772},
  year={2024}
}

@article{han2024wildguard,
  title={Wildguard: Open one-stop moderation tools for safety risks, jailbreaks, and refusals of llms},
  author={Han, Seungju and Rao, Kavel and Ettinger, Allyson and Jiang, Liwei and Lin, Bill Yuchen and Lambert, Nathan and Choi, Yejin and Dziri, Nouha},
  journal={Advances in Neural Information Processing Systems},
  volume={37},
  pages={8093--8131},
  year={2024}
}

@article{bianchi2023safety,
  title={Safety-tuned llamas: Lessons from improving the safety of large language models that follow instructions},
  author={Bianchi, Federico and Suzgun, Mirac and Attanasio, Giuseppe and R{\"o}ttger, Paul and Jurafsky, Dan and Hashimoto, Tatsunori and Zou, James},
  journal={arXiv preprint arXiv:2309.07875},
  year={2023}
}

@article{tuan2024towards,
  title={Towards safety and helpfulness balanced responses via controllable large language models},
  author={Tuan, Yi-Lin and Chen, Xilun and Smith, Eric Michael and Martin, Louis and Batra, Soumya and Celikyilmaz, Asli and Wang, William Yang and Bikel, Daniel M},
  journal={arXiv preprint arXiv:2404.01295},
  year={2024}
}

@inproceedings{rottger-etal-2024-xstest,
    title = "{XST}est: A Test Suite for Identifying Exaggerated Safety Behaviours in Large Language Models",
    author = {R{\"o}ttger, Paul  and
      Kirk, Hannah  and
      Vidgen, Bertie  and
      Attanasio, Giuseppe  and
      Bianchi, Federico  and
      Hovy, Dirk},
    editor = "Duh, Kevin  and
      Gomez, Helena  and
      Bethard, Steven",
    booktitle = "Proceedings of the 2024 Conference of the North American Chapter of the Association for Computational Linguistics: Human Language Technologies (Volume 1: Long Papers)",
    month = jun,
    year = "2024",
    address = "Mexico City, Mexico",
    publisher = "Association for Computational Linguistics",
    url = "https://aclanthology.org/2024.naacl-long.301/",
    doi = "10.18653/v1/2024.naacl-long.301",
    pages = "5377--5400"
}

@inproceedings{shi-etal-2024-navigating,
    title = "Navigating the {O}ver{K}ill in Large Language Models",
    author = "Shi, Chenyu  and
      Wang, Xiao  and
      Ge, Qiming  and
      Gao, Songyang  and
      Yang, Xianjun  and
      Gui, Tao  and
      Zhang, Qi  and
      Huang, Xuanjing  and
      Zhao, Xun  and
      Lin, Dahua",
    editor = "Ku, Lun-Wei  and
      Martins, Andre  and
      Srikumar, Vivek",
    booktitle = "Proceedings of the 62nd Annual Meeting of the Association for Computational Linguistics (Volume 1: Long Papers)",
    month = aug,
    year = "2024",
    address = "Bangkok, Thailand",
    publisher = "Association for Computational Linguistics",
    url = "https://aclanthology.org/2024.acl-long.253/",
    doi = "10.18653/v1/2024.acl-long.253",
    pages = "4602--4614"
}

@article{an2024automatic,
  title={Automatic pseudo-harmful prompt generation for evaluating false refusals in large language models},
  author={An, Bang and Zhu, Sicheng and Zhang, Ruiyi and Panaitescu-Liess, Michael-Andrei and Xu, Yuancheng and Huang, Furong},
  journal={arXiv preprint arXiv:2409.00598},
  year={2024}
}

@article{cui2024or,
  title={Or-bench: An over-refusal benchmark for large language models},
  author={Cui, Justin and Chiang, Wei-Lin and Stoica, Ion and Hsieh, Cho-Jui},
  journal={arXiv preprint arXiv:2405.20947},
  year={2024}
}

@article{zhang2025falsereject,
  title={Falsereject: A resource for improving contextual safety and mitigating over-refusals in llms via structured reasoning},
  author={Zhang, Zhehao and Xu, Weijie and Wu, Fanyou and Reddy, Chandan K},
  journal={arXiv preprint arXiv:2505.08054},
  year={2025}
}

@article{perez2022red,
  title={Red teaming language models with language models},
  author={Perez, Ethan and Huang, Saffron and Song, Francis and Cai, Trevor and Ring, Roman and Aslanides, John and Glaese, Amelia and McAleese, Nat and Irving, Geoffrey},
  journal={arXiv preprint arXiv:2202.03286},
  year={2022}
}

@article{hong2024curiosity,
  title={Curiosity-driven red-teaming for large language models},
  author={Hong, Zhang-Wei and Shenfeld, Idan and Wang, Tsun-Hsuan and Chuang, Yung-Sung and Pareja, Aldo and Glass, James and Srivastava, Akash and Agrawal, Pulkit},
  journal={arXiv preprint arXiv:2402.19464},
  year={2024}
}

@article{zou2023universal,
  title={Universal and transferable adversarial attacks on aligned language models},
  author={Zou, Andy and Wang, Zifan and Carlini, Nicholas and Nasr, Milad and Kolter, J Zico and Fredrikson, Matt},
  journal={arXiv preprint arXiv:2307.15043},
  year={2023}
}

@article{liu2023jailbreaking,
  title={Jailbreaking chatgpt via prompt engineering: An empirical study},
  author={Liu, Yi and Deng, Gelei and Xu, Zhengzi and Li, Yuekang and Zheng, Yaowen and Zhang, Ying and Zhao, Lida and Zhang, Tianwei and Wang, Kailong and Liu, Yang},
  journal={arXiv preprint arXiv:2305.13860},
  year={2023}
}

@article{lapid2024open,
  title={Open sesame! universal black-box jailbreaking of large language models},
  author={Lapid, Raz and Langberg, Ron and Sipper, Moshe},
  journal={Applied Sciences},
  volume={14},
  number={16},
  pages={7150},
  year={2024},
  publisher={MDPI}
}

@INPROCEEDINGS {10992337,
author = { Chao, Patrick and Robey, Alexander and Dobriban, Edgar and Hassani, Hamed and Pappas, George J. and Wong, Eric },
booktitle = { 2025 IEEE Conference on Secure and Trustworthy Machine Learning (SaTML) },
title = {{ Jailbreaking Black Box Large Language Models in Twenty Queries }},
year = {2025},
volume = {},
ISSN = {},
pages = {23-42},
doi = {10.1109/SaTML64287.2025.00010},
url = {https://doi.ieeecomputersociety.org/10.1109/SaTML64287.2025.00010},
publisher = {IEEE Computer Society},
address = {Los Alamitos, CA, USA},
month =apr}

@inproceedings{zeng-etal-2024-johnny,
    title = "How Johnny Can Persuade {LLM}s to Jailbreak Them: Rethinking Persuasion to Challenge {AI} Safety by Humanizing {LLM}s",
    author = "Zeng, Yi  and
      Lin, Hongpeng  and
      Zhang, Jingwen  and
      Yang, Diyi  and
      Jia, Ruoxi  and
      Shi, Weiyan",
    editor = "Ku, Lun-Wei  and
      Martins, Andre  and
      Srikumar, Vivek",
    booktitle = "Proceedings of the 62nd Annual Meeting of the Association for Computational Linguistics (Volume 1: Long Papers)",
    month = aug,
    year = "2024",
    address = "Bangkok, Thailand",
    publisher = "Association for Computational Linguistics",
    url = "https://aclanthology.org/2024.acl-long.773/",
    doi = "10.18653/v1/2024.acl-long.773",
    pages = "14322--14350"
}

@article{andriushchenko2024jailbreaking,
      title={Jailbreaking Leading Safety-Aligned LLMs with Simple Adaptive Attacks}, 
      author={Andriushchenko, Maksym and Croce, Francesco and Flammarion, Nicolas},
      journal={arXiv preprint arXiv:2404.02151},
      year={2024}
}

@article{robey2023smoothllm,
  title={Smoothllm: Defending large language models against jailbreaking attacks},
  author={Robey, Alexander and Wong, Eric and Hassani, Hamed and Pappas, George J},
  journal={arXiv preprint arXiv:2310.03684},
  year={2023}
}

@article{wei2023jailbreak,
  title={Jailbreak and guard aligned language models with only few in-context demonstrations},
  author={Wei, Zeming and Wang, Yifei and Li, Ang and Mo, Yichuan and Wang, Yisen},
  journal={arXiv preprint arXiv:2310.06387},
  year={2023}
}

@article{wang2024defending,
  title={Defending llms against jailbreaking attacks via backtranslation},
  author={Wang, Yihan and Shi, Zhouxing and Bai, Andrew and Hsieh, Cho-Jui},
  journal={arXiv preprint arXiv:2402.16459},
  year={2024}
}

@article{Xie2023DefendingCA,
  title={Defending ChatGPT against jailbreak attack via self-reminders},
  author={Yueqi Xie and Jingwei Yi and Jiawei Shao and Justin Curl and Lingjuan Lyu and Qifeng Chen and Xing Xie and Fangzhao Wu},
  journal={Nature Machine Intelligence},
  year={2023},
  volume={5},
  pages={1486-1496},
}

@inproceedings{mai2025you,
  title={You Can't Eat Your Cake and Have It Too: The Performance Degradation of LLMs with Jailbreak Defense},
  author={Mai, Wuyuao and Hong, Geng and Chen, Pei and Pan, Xudong and Liu, Baojun and Zhang, Yuan and Duan, Haixin and Yang, Min},
  booktitle={Proceedings of the ACM on Web Conference 2025},
  pages={872--883},
  year={2025}
}

@article{borah2025alignment,
  title={Alignment Quality Index (AQI): Beyond Refusals: AQI as an Intrinsic Alignment Diagnostic via Latent Geometry, Cluster Divergence, and Layer wise Pooled Representations},
  author={Borah, Abhilekh and Sharma, Chhavi and Khanna, Danush and Bhatt, Utkarsh and Singh, Gurpreet and Abdullah, Hasnat Md and Ravi, Raghav Kaushik and Jain, Vinija and Patel, Jyoti and Singh, Shubham and others},
  journal={arXiv preprint arXiv:2506.13901},
  year={2025}
}

\clearpage
\onecolumn

\section*{Supplementary Material}

\subsection*{A. Reproducibility Overview}

This supplement provides implementation details for constructing ParaGuard and computing the semantic confusion metrics. The goal is to make the dataset construction, refusal labeling, neighbor retrieval, and CI/CR/CD computation easier to reproduce. Code and data will be released upon acceptance.

\subsection*{B. ParaGuard Construction}

\begin{algorithm}[h]
\caption{ParaGuard Construction}
\label{alg:paraguard}
\footnotesize
\begin{algorithmic}[1]
\REQUIRE Source benchmarks $\mathcal{D}=\{\text{OR-Bench}, \text{USEBench}, \text{PHTest}\}$; seed count $N$; variants per seed $M=5$
\ENSURE Filtered paraphrase corpus $\mathcal{P}$

\STATE Load prompts from all source benchmarks and normalize them into a common schema.
\STATE Remove empty prompts and exact duplicates.
\STATE Sample $N$ seed prompts using a fixed random seed.
\STATE Initialize $\mathcal{P}\leftarrow \emptyset$.

\FOR{each seed prompt $s$}
    \STATE Generate $M$ candidate paraphrases using lexical rewording, register shift, keyword softening/hardening, and intent-preserving rewrites.
    \FOR{each candidate paraphrase $v$}
        \STATE Compute sentence-embedding similarity $\mathrm{sim}(s,v)$.
        \STATE Compute character-level lexical overlap $\mathrm{overlap}(s,v)$.
        \STATE Compute ensemble safety risk score $\mathrm{risk}(v)$.
        \IF{$\mathrm{sim}(s,v)\geq 0.60$ and $\mathrm{overlap}(s,v)\leq 0.90$ and $\mathrm{risk}(v)$ passes the safety gate}
            \STATE Add $v$ to $\mathcal{P}$ with its seed ID, rewrite type, similarity, overlap, and risk metadata.
        \ENDIF
    \ENDFOR
\ENDFOR

\STATE \textbf{return} $\mathcal{P}$.
\end{algorithmic}
\end{algorithm}

\subsection*{C. Refusal Labeling, Retrieval, and Confusion Scoring}

\begin{algorithm}[h]
\caption{Semantic Confusion Measurement}
\label{alg:confusion-pipeline}
\footnotesize
\begin{algorithmic}[1]
\REQUIRE ParaGuard corpus $\mathcal{P}$; target model $f$; sentence encoder $E$; neighbor count $k$; weights $(w_d,w_p,w_\pi)$; threshold $\tau$
\ENSURE Dataset-level CI, CR@$\,\tau$, and CD

\STATE Initialize accepted set $\mathcal{A}\leftarrow \emptyset$ and rejected set $\mathcal{R}\leftarrow \emptyset$.

\FOR{each prompt $x\in\mathcal{P}$}
    \STATE Query $f(x)$ using deterministic decoding.
    \STATE Label the response as \texttt{ACCEPT} or \texttt{REJECT} using explicit refusal cues.
    \IF{$x$ is labeled \texttt{ACCEPT}}
        \STATE Add $x$ to $\mathcal{A}$.
    \ELSIF{$x$ is labeled \texttt{REJECT}}
        \STATE Add $x$ to $\mathcal{R}$.
    \ENDIF
\ENDFOR

\STATE Encode all prompts in $\mathcal{A}$ using $E$ and build a FAISS index.
\STATE Initialize confusion list $S\leftarrow \emptyset$.

\FOR{each rejected prompt $r\in\mathcal{R}$}
    \STATE Retrieve the top-$k$ nearest accepted prompts $\mathcal{N}_k(r)$ from the FAISS index.
    \STATE Initialize pair-score list $S_r\leftarrow \emptyset$.
    \FOR{each accepted neighbor $a\in\mathcal{N}_k(r)$}
        \STATE Compute token drift $\mathrm{Drift}(a,r)$.
        \STATE Compute token probability shift $\mathrm{ProbShift}(a,r)$.
        \STATE Compute perplexity contrast $\Delta\mathrm{PPL}(a,r)$.
        \STATE Compute $\mathrm{CS}(a,r)=w_d\mathrm{Drift}(a,r)+w_p\mathrm{ProbShift}(a,r)+w_\pi\Delta\mathrm{PPL}(a,r)$.
        \STATE Add $\mathrm{CS}(a,r)$ to $S_r$.
    \ENDFOR
    \STATE Compute $\mathrm{CI}(r)=\frac{1}{k}\sum_{a\in\mathcal{N}_k(r)}\mathrm{CS}(a,r)$.
    \STATE Add $\mathrm{CI}(r)$ to $S$.
\ENDFOR

\STATE Compute $\mathrm{CI}=\frac{1}{|\mathcal{R}|}\sum_{r\in\mathcal{R}}\mathrm{CI}(r)$.
\STATE Compute $\mathrm{CR}@\tau=\frac{1}{|\mathcal{R}|}\sum_{r\in\mathcal{R}}\mathbf{1}\{\mathrm{CI}(r)\geq\tau\}$.
\STATE Compute $\mathrm{CD}=\operatorname{stdev}\left(\{\mathrm{CI}(r)\}_{r\in\mathcal{R}}\right)$.
\STATE \textbf{return} CI, CR@$\,\tau$, and CD.
\end{algorithmic}
\end{algorithm}

\subsection*{D. Main Reproducibility Settings}

\begin{table}[h]
\centering
\small
\caption{Main implementation settings used in the experiments.}
\label{tab:supp-settings}
\begin{tabular}{ll}
\toprule
Setting & Value \\
\midrule
Source datasets & OR-Bench, USEBench, PHTest \\
Number of sampled seeds & 2,000 \\
Variants per seed & 5 \\
Sentence encoder & MiniLM-L6-v2 \\
Similarity gate & $\geq 0.60$ \\
Lexical overlap gate & $\leq 0.90$ \\
Neighbor count & $k=5$ \\
Metric weights & $(w_d,w_p,w_\pi)=(0.4,0.1,0.5)$ \\
Confusion threshold & $\tau=0.75$ \\
Decoding temperature & 0 \\
\bottomrule
\end{tabular}
\end{table}

\subsection*{E. Refusal Labeling Details}

Model responses are labeled using explicit refusal cues. A response is marked as \texttt{REJECT} if it contains refusal expressions such as ``sorry'', ``cannot'', ``can't'', ``not allowed'', ``I cannot help'', and many more similar safety-oriented refusal phrases. Otherwise, it is marked as \texttt{ACCEPT}. Failed or ambiguous model calls should be excluded from metric computation and reported separately.

\subsection*{F. Notes on Reproducibility}

All random sampling steps use a fixed random seed. Model outputs may still vary across model versions, API backends, tokenizer implementations, and decoding infrastructure. For this reason, we report model names, decoding settings, filtering thresholds, neighbor count, metric weights, and the confusion threshold. The released code will include the final dataset, evaluation scripts, and figure-generation utilities.

\end{document}